%% file: egpaper_final.tex
\documentclass[10pt,twocolumn,letterpaper]{article}

\usepackage{3dv}
\usepackage{times}
\usepackage{epsfig}
\usepackage{graphicx}
\usepackage{amsmath}
\usepackage{amssymb}
\usepackage{multirow}
\usepackage{enumitem}
\usepackage{booktabs}
\usepackage{hyperref}

\def\ie{\textit{i.e.}}

\def\etal{\textit{et al.}}

\newcommand{\printfnsymbol}[1]{%
  \textsuperscript{\@fnsymbol{#1}}%
}
\newcommand{\vect}[1]{\mathbf{#1}}

\threedvfinalcopy 

\def\threedvPaperID{****} 

\ifthreedvfinal\fi
\begin{document}
\title{Unsupervised Adversarial Depth Estimation using Cycled Generative Networks}

\author{Andrea Pilzer$^\dagger$\thanks{The authors contributed equally in this work.} \quad  Dan Xu$^\ddagger$\footnotemark[1] \quad  Mihai Marian Puscas$^\dagger$\footnotemark[1] \quad Elisa Ricci$^{\dagger \diamond}$ \quad  Nicu Sebe$^\dagger$\\
	$^\dagger$DISI, University of Trento, via Sommarive 14, Povo TN, Italy\\
	$^\ddagger$Department of Engineering Science, University of Oxford, 17 Parks Road, Oxford, UK\\
	$^\diamond$Technologies of Vision, Fondazione Bruno Kessler, via Sommarive 18, Povo TN, Italy\\
	{\tt\small \{andrea.pilzer, mihaimarian.puscas, e.ricci, niculae.sebe\}@unitn.it, danxu@robots.ox.ac.uk}
}

\maketitle

\begin{abstract}
	While recent deep monocular depth estimation approaches based on supervised regression have achieved remarkable performance, costly ground truth annotations are required during training. To cope with this issue, in this paper we present a novel unsupervised deep learning approach for predicting depth maps and show that the depth estimation task can be effectively tackled within an adversarial learning framework. Specifically, we propose a deep generative network that learns to predict the correspondence field (\ie~the disparity map) between two image views in a calibrated stereo camera setting. The proposed architecture consists of two generative sub-networks
	jointly trained with adversarial learning for reconstructing the disparity map and organized in a cycle such as to provide mutual constraints and supervision to each other. Extensive experiments on the publicly available datasets KITTI and Cityscapes demonstrate the effectiveness of the proposed model and competitive results with state of the art methods. The code and trained model are available: \url{https://github.com/andrea-pilzer/unsup-stereo-depthGAN}
\end{abstract}
\vspace{-18pt}

\input{intro.tex}
\input{related.tex}
\input{method.tex}

\input{exp.tex}

\section{Conclusion}
We have presented a novel approach for unsupervised deep learning for the depth estimation task using the adversarial learning strategy in a proposed cycled generative network structure. The new approach provides a new insight to the community that shows depth estimation can be effectively tackled via an unsupervised adversarial learning of the stereo image synthesis. More specifically, a generative deep network model is proposed to learn to predict the disparity map between two image views under a calibrated stereo camera setting. Two symmetric generative sub-networks are respectively designed to generate images from different views, and they are further merged to form a closed cycle which is able to provide strong constraint and supervision to optimize better the dual generators of the two sub-networks. Extensive experiments are conducted on two publicly available datasets (\ie~KITTI and Cityscapes). The results demonstrate the effectiveness of the proposed model, and show very competitive performance compared to state-of-the-arts on the KITTI dataset.
\par The future work would contain using attention mechanism to guide the learning of the feature representations of the generators, and also consider using the graphical models for structured prediction on the output disparity map to have predictions with better scene structures.

{\small
\bibliographystyle{ieee}
\bibliography{egbib}
}

\end{document}

%% file: intro.tex
\section{Introduction}
\label{sec:intro}
As one of the fundamental problems in computer vision, depth estimation has received a substantial interest in the past, also motivated by its importance in 
various application scenarios, such as robotics navigation, 3D reconstruction, virtual reality and autonomous driving. Over the last few years
the performances of depth estimation methods have been significantly improved thanks to advanced deep learning techniques. 

\begin{figure}[!t]
	\includegraphics[width=\columnwidth]{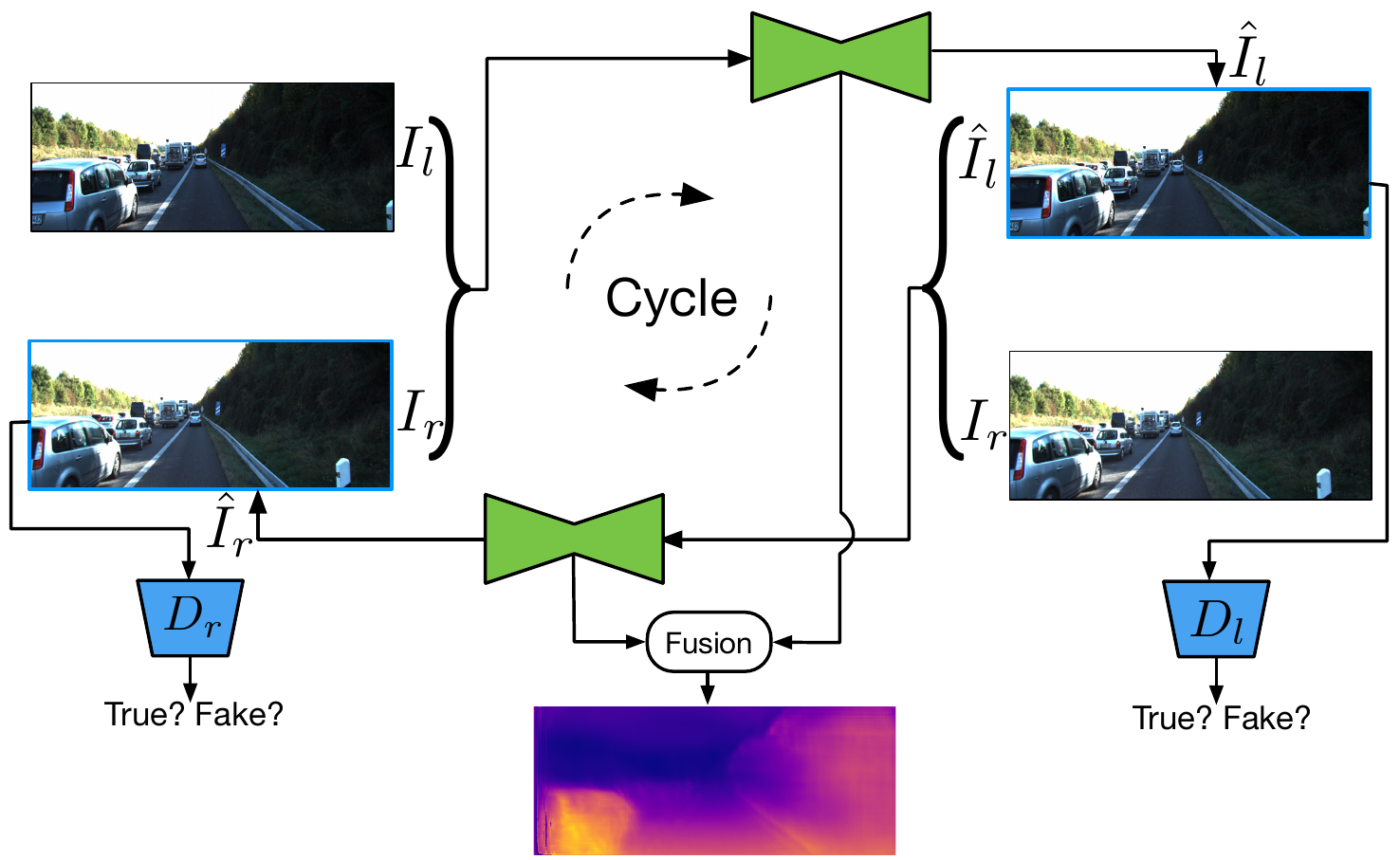}
	\caption{Motivation of the proposed unsupervised depth estimation approach using cycled generative networks optimized with adversarial learning. The left and right image synthesis in a cycle provides each other strong constraint and supervision to better optimize both generators. The $\vect{\hat{I}}_r$ and $\vect{\hat{I}}_l$ are synthesized images. Final depth estimation is obtained by fusing the output from both generators.}
	\label{fig:motivation}
	\vspace{-15pt}
\end{figure}

Most previous works considering deep architectures for predicting depth maps operate in a supervised learning setting~\cite{eigen2014depth,ladicky2014pulling,liu2016learning,xu2018monocular} and, 
specifically, devise powerful deep regression models with Convolutional Neural Networks (CNN). These models are used for monocular depth estimation, \textit{i.e.} they 
are trained to learn  the transformation from the RGB image domain to the depth domain in a pixel-to-pixel fashion. 
In this context, multi-scale CNN models have shown to be especially effective for estimating depth maps~\cite{eigen2014depth}. Upon these,
probabilistic graphical models, such as Conditional Random Fields (CRFs), implemented as neural networks for end-to-end optimization, have 
proved to be beneficial, boosting the performance of deep regression models \cite{liu2016learning,xu2018monocular}. 
However, supervised learning models require ground-truth depth data which are usually costly to acquire. This problem is especially relevant
with deep learning architectures, as large amount of data are typically required to produce satisfactory performance. Furthermore, supervised 
monocular depth estimation can be regarded as an ill-posed problem due to the scale ambiguity issue~\cite{saxena2006learning}.  


To tackle these problems, recently unsupervised learning-based approaches for depth estimation have been introduced~\cite{luo2016efficient,mayer2016large}.
These methods operate by learning the correspondence field (\ie~the disparity map) between the two different image views of a calibrated stereo camera using 
only the rectified left and right images. Then, given several camera parameters, the depth maps can be calculated using the predicted disparity maps. Significant progresses 
have been made along this research line~\cite{garg2016unsupervised,godard2017unsupervised,wang2017learning}. In particular, Godard~\etal~\cite{godard2017unsupervised} 
proposed to estimate both the direct and the reverse disparity maps using a single generative network and utilized the consistency between left and right disparity maps 
to constrain 
on the model learning. Other works proposed to facilitate the depth 
estimation by jointly learning the camera pose~\cite{zhou2017unsupervised,mahjourian2018unsupervised}. These works optimized their models  relying on the 
supervision from the 
image synthesis of an expected view, whose quality plays a direct influence on the performance of the estimated disparity map. However, all of these works 
only considered a reconstruction loss and none of them have explored using adversarial learning to improve the generation of the synthesized images.

In this paper, we follow the unsupervised learning setting and propose a novel end-to-end trainable deep network model for adversarial learning-based depth estimation given
stereo image pairs.
The proposed approach consists of two generative sub-networks which predict the disparity map from the left to the right view and viceversa. 
The two sub-networks are organized in a cycle (Fig.~\ref{fig:motivation}), such as to perform the image synthesis of different views in a closed loop. This new 
network design provides strong constraint and supervision for each image view, facilitating the optimization of both generators from the two sub-networks 
which are jointly learned with an adversarial learning strategy. The final disparity map is produced by combining the output from the two generators. 

In summary, the main contributions of this paper are threefolds: 
\vspace{-9pt} 


\begin{itemize}[leftmargin=*]
 \setlength\itemsep{-0.3em}
\item To the best of our knowledge, we are the first to explore using adversarial learning to facilitate the image synthesis of different views 
in a unified deep network for improving the unsupervised depth estimation;
\item We present a new cycled generative network structure for unsupervised depth estimation which can learn both the forward and the reverse 
disparity maps, and can synthesize the different image views in a closed loop. Compared with the existing generative network structures, the proposed cycled generative 
network is able to enforce stronger constraints from each image view and better optimize the network generators.
\item Extensive experiments on two large publicly available datasets (\ie~KITTI and Cityscapes) demonstrate the effectiveness of both the adversarial image synthesis 
and the cycled generative network structure. 
\end{itemize}

%% file: related.tex
\section{Related Work}
\label{sec:related}
\textbf{Supervised Depth Estimation.} Supervised deep learning greatly improved the performance of depth estimation. Given enough ground-truth depth training data, deep neural networks based approaches have achieved very promising performances in recent years. Multiple large-scale depth-contained datasets~\cite{Silberman:ECCV12,saxena2009make3d,kitti,Cityscapes} have been published. In a single view setting, NYUD~\cite{Silberman:ECCV12} presents indoor images while Make3D~\cite{saxena2009make3d} is recorded in outdoors. Instead KITTI~\cite{kitti} and Cityscapes~\cite{Cityscapes} are collected in outdoors with calibrated stereo cameras. Based on these datasets, a significant effort has been made for the supervised monocular depth estimation task~\cite{eigen2014depth,liu2016learning,zhuo2015indoor,laina2016deeper,xu2018monocular}. The multi-scale CNN~\cite{eigen2014depth} and probabilistic graphical models based deep networks~\cite{liu2016learning,xu2018monocular,xu2017multi} also show an obvious performance boosting on the task. Xu~\etal~\cite{xu2018structured} first introduce a structured attention mechanism for learning better multi-scale deep representations for the task. However, the supervised-based approaches rely on the expensive ground-truth depth data during training, which are not flexible to deploy crossing application scenarios. 

\label{sec:method}
\begin{figure*}
	\begin{center}
		\includegraphics[width=0.95\textwidth]{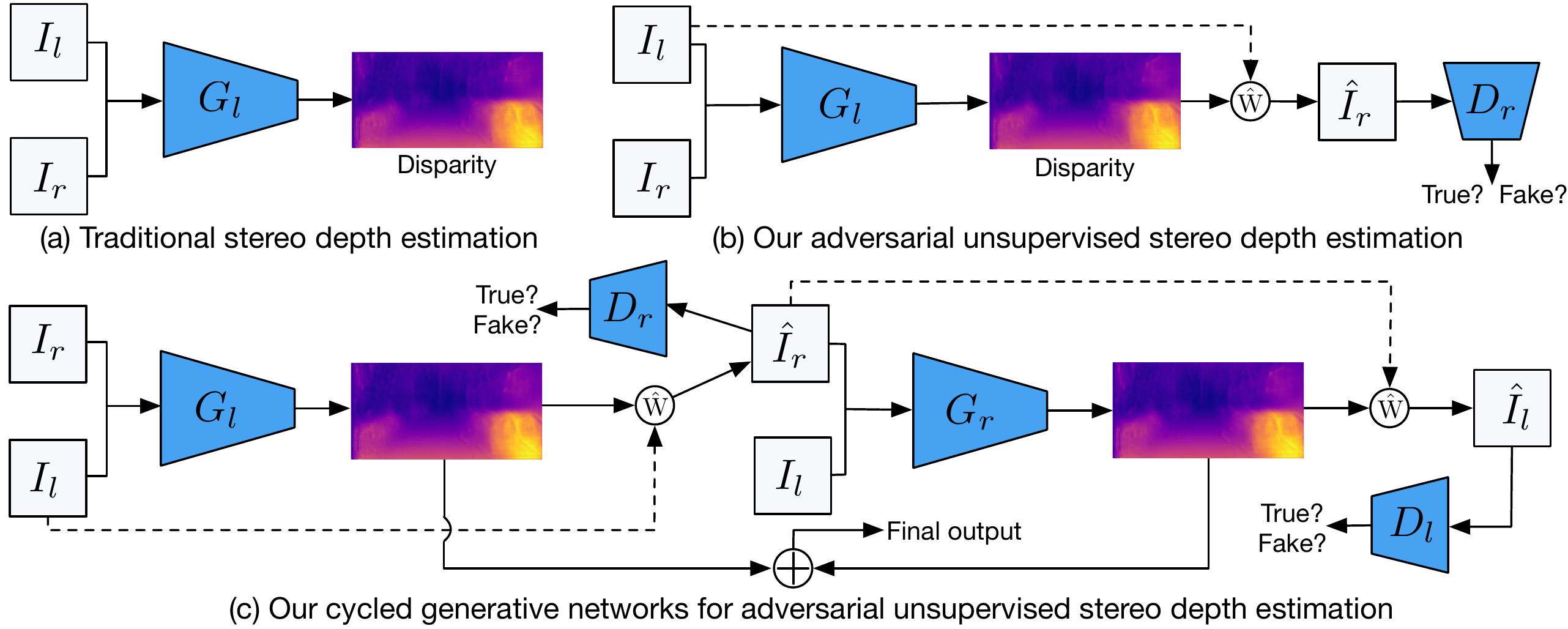}
		\caption{An illustrative comparison of different methods for unsupervised stereo depth estimation: (a) traditional stereo-matching-based depth estimation, (b) the proposed unsupervised adversarial depth estimation and (c) the proposed cycled generative networks for unsupervised adversarial depth estimation. The symbols $D_l$, $D_r$ denote discriminators, and $G_l$, $G_r$ denote generators. The symbol $\hat{\mathrm{W}}$ denotes a warping operation.}
	\end{center}
	\label{fig:comparison}
\vspace{-15pt}
\end{figure*}

\textbf{Unsupervised Depth Estimation.} A more recent trend is unsupervised-based depth estimation~\cite{kuznietsov2017semi,mahjourian2018unsupervised,wang2017learning,zhan2018unsupervised}. A remarkable advantage of unsupervised estimation lies in avoiding the use of costly ground truth depth annotations in training. Deep stereo matching models~\cite{luo2016efficient,mayer2016large} are proposed for direct disparity estimation. In an indirect means, Garg~\etal~\cite{garg2016unsupervised} propose a classic approach for unsupervised monocular depth estimation based on image synthesis. Godard~\etal~\cite{godard2017unsupervised} propose to use forward and backward reconstructions of the different image views, and multiple optimization losses are considered in the model. Zhou~\etal~\cite{zhou2017unsupervised} jointly learn the depth and the camera pose as a reinforcement in a single deep network. There are also works jointly learning the scene depth and ego-motion in monocular videos without using groundtruth data~\cite{wang2018learning,yang2018every}. However, none of these works considers the adversarial learning scheme in their models to improve the image generation quality for better depth estimation.

\textbf{GANs.} Generative-adversarial networks (GANs) have attracted a lot of attention for its advantage in generation problems. Godfellow~\etal~\cite{goodfellow2014generative} revisit the generative adversarial learning strategy and show interesting results in the image generation task. After that, GANs are applied into various generation applications, and different GAN models are developed, such as CycleGAN~\cite{CycleGAN2017} and DualGAN~\cite{yi2017dualgan}. There are few works in the literature considering GAN models for the more challenging depth estimation task. Although Kundu~\etal~\cite{adadepth} investigate adversarial learning for the task, they utilize it in a context of domain adaptation in a single-track network, using a semi-supervised setting with an extra synthetic dataset, while ours considers a fully unsupervised setting and the adversarial learning in a cycled generative network aims to help the reconstruction of better image views. Both the intuition and the network design are significantly different.

%% file: method.tex
\section{The Proposed Approach}

We propose a novel approach for unsupervised adversarial depth estimation using cycled generative networks. An illustrative comparison of different unsupervised depth estimation models is shown in Fig.~\ref{fig:comparison}. Fig.~\ref{fig:comparison}a shows traditional stereo matching based depth estimation approaches, which basically learn a stereo matching network for directly predicting the disparity~\cite{luo2016efficient}. Different from the traditional stereo approaches, we estimate the disparity in an indirect means through image synthesis from different views with the adversarial learning strategy as shown in Fig.~\ref{fig:comparison}b.  Fig.~\ref{fig:comparison}c shows our full model using the proposed cycled generative networks for the task. In this section we first give the problem statement, and then present the proposed adversarial learning-based unsupervised stereo depth estimation, and finally we illustrate the proposed full model and introduce the overall end-to-end optimization objective and the testing process.


\begin{figure*}[!t]
\begin{center}
\includegraphics[width=0.95\textwidth]{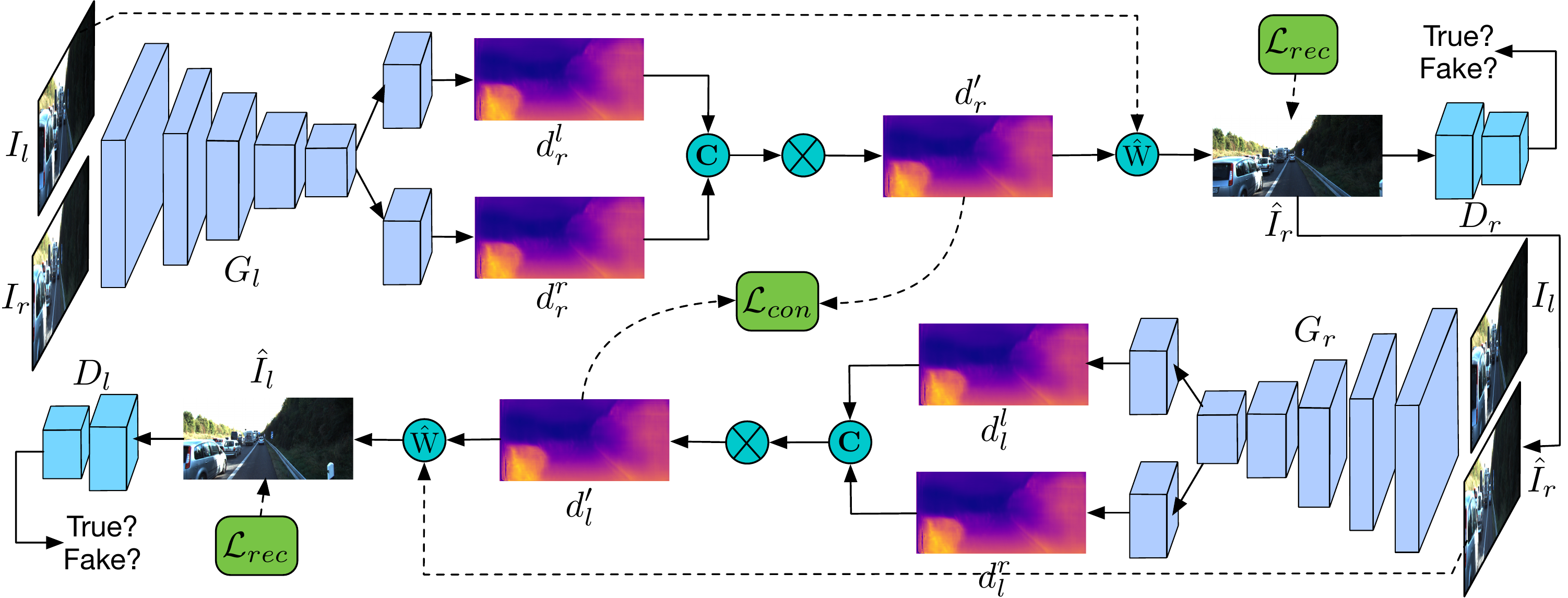}
\caption{Illustration of the detailed framework of the proposed cycled generative networks for unsupervised adversarial depth estimation. The symbol $\copyright$ denotes a concatenation operation; $\mathcal{L}_{rec}$ represents the reconstruction loss for different generators; $\mathcal{L}_{con}$ denotes a consistence loss between the disparity maps generated from the two generators.}
\end{center}
\label{fig:detailedgan}
\vspace{-20pt}
\end{figure*}

\subsection{Problem Statement}
We target at estimating a disparity map given a pair of images from a calibrated stereo camera. The problem can be formally defined as follows: given a left image $\vect{I}_l$ and a right image $\vect{I}_r$ from the camera, we are interested in predicting a disparity map $\vect{d}$ in which each pixel value represents an offset of the corresponding pixel between the left and the right image. If given the baseline distance $b_d$ between the left and the right camera and the camera focal length $f_l$, a depth map $\vect{D}$ can be calculated with the formula of $\vect{D}=(b_d * f_l )/ \vect{d}$. We indirectly learn the disparity through the image synthesis. Specifically, assume that a left-to-right disparity $\vect{d}_r^{(l)}$ is produced from a generative network $G_l$ with the left-view image $\vect{I}_l$ as input, and then a warping function $f_w(\cdot)$ is used to perform the synthesis of the right image view by sampling from $\vect{I}_l$, \ie~$\hat{\vect{I}}_r = f_w(\vect{d}_r^{(l)}, \vect{I}_l)$. A reconstruction loss between $\hat{\vect{I}}_r$ and $\vect{I}_r$ is thus utilized to provide supervision in optimizing the network $G_l$. 

\subsection{Unsupervised Adversarial Depth Estimation}
\label{half-cycle}
We now introduce the proposed unsupervised adversarial depth estimation approach. Assuming we have a generative network $G_l$ composed of two sub-networks, a generative sub-network $G_l^{(l)}$ with input $\vect{I}_l$ and a generative sub-network $G_l^{(r)}$ with input $\vect{I}_r$. These are used to produce two distinct left-to-right disparity maps $\vect{d}_r^{(l)}$ and $\vect{d}_r^{(r)}$ respectively, \ie~$\vect{d}_r^{(l)} = G_l^{(l)}(\vect{I}_l)$ and $\vect{d}_r^{(r)} = G_l^{(r)}(\vect{I}_r)$. The sub-network $G_l^{(l)}$ and $G_l^{(r)}$ exploit the same network structure using a convolutional encoder-decoder, where the encoders aim at obtaining compact image representations and could be shared to reduce the network capacity. Since the two disparity maps are produced from different  input images, and show complementary characteristics, they are fused using a linear combination implemented as concatenation and $1\times 1$ convolution, and we obtain an enhanced disparity map $\vect{d}'_r$, which is used to synthesize a right view image $\hat{\vect{I}}_r$ via the warping operation, \ie~$\hat{\vect{I}}_r = f_w(\vect{d}'_r, \vect{I}_l)$. Then we use an $L1$-norm reconstruction loss $\mathcal{L}_{rec}$ for optimization as follows:
\begin{equation}
 \mathcal{L}_{rec}^{(r)} = \lVert \vect{I}_r - f_w(\vect{d}'_r, \vect{I}_l) \lVert_1
 \end{equation} 
 
To improve the generation quality of the image $\hat{\vect{I}}_r$ and benefit from the advantage of adversarial learning, we propose to use adversarial learning here for a better optimization due to its demonstrated powerful ability in the image generation task~\cite{goodfellow2014generative}. For the synthesized image $\hat{\vect{I}}_r$, a discriminators $D_r$ outputting a scalar value which is used to discriminate if the image $\hat{\vect{I}}_r$ or $\vect{I}_r$ is fake or true, and thus the adversarial objective for the generative network can be formulated as follows:
\begin{equation}
\begin{split}
  \hspace*{-0.3cm} \mathcal{L}_{gan}^{(r)}(G_l, & D_r, \vect{I}_{l}, \vect{I}_{r}) = 
  \mathbb{E}_{\vect{I}_r \sim p(\vect{I}_r)}[\log D_r(\vect{I}_r)]  \\ &+  \mathbb{E}_{\vect{I}_l \sim p(\vect{I}_l)}[\log(1 - D_r(f_w(\vect{d}'_r, \vect{I}_l)))]
\end{split}
\end{equation}
where we adopt a cross-entropy loss to measure the expectation of the image $\vect{I}_l$ and $\vect{I}_r$ against the distribution of the left and the right view images $p(\vect{I}_l)$ and $p(\vect{I}_r)$ respectively. Then the joint optimization loss is the combination of the reconstruction loss and the adversarial loss written as:
\begin{equation}
\mathcal{L}_o^{(r)} = \gamma_1 \mathcal{L}_{rec}^{(r)} +  \gamma_2 \mathcal{L}_{gan}^{(r)}
\end{equation}
where $\gamma_1$ and $\gamma_2$ are the weights for balancing the loss magnitude of the two parts to stabilize the training process. In the testing phase, the inferred $\vect{d}'_r$ is the final output.

\subsection{Cycled Generative Networks for Adversarial Depth Estimation}

In the previous section, we presented the adversarial learning-based depth estimation approach which reconstructs from one image view to the other one in a straightforward way. In order to make the image reconstruction from different views implicitly constrain on each other, we further propose a cycled generative network structure. An overview of the proposed network structure is shown in Fig.~\ref{sec:method}. The network produces two distinct disparity maps from different view directions, and synthesizes different-view images in a closed loop. In our network design, not only the different view reconstruction loss helps for better optimization of the generators, but also the two disparity maps are connected with a consistence loss to provide strong supervision from each half cycle. 

We described the half-cycle generative network with adversarial learning in Section~\ref{half-cycle}. The cycled generative network is based on the half-cycle structure. To simplify the description, we follow the notations used in Section~\ref{half-cycle}. Assume we have obtained a synthesized image $\hat{\vect{I}}_r$ from the half-cycle network, and then $\hat{\vect{I}}_r$ is further used as input of the next cycle generative network. Let us denote the generator as $G_r$, which we exploit the encoder-decoder network structure similar as $G_l$ in Sec.~\ref{half-cycle}. The encoder part of $G_r$ can be also shared with the encoder of $G_l$ to have a more compact network model (we show the performance difference between using and not using the sharing scheme), and the two distinct decoders are used to produce two right-to-left disparity maps $\vect{d}_l^{(l)}$ and $\vect{d}_l^{(r)}$ corresponding the left- and the right-view input images respectively. The two maps are also combined with the combination and the convolution operation to have a fused disparity map $\vect{d}'_l$. Then we synthesize the left-view image $\hat{\vect{I}}_l$ via the warping operation as $\hat{\vect{I}}_l = f_w(\vect{d}'_l, \vect{I}_r)$. An $L1$-norm reconstruction loss is used for optimizing the generator $G_r$. Then the objective for optimizing the two generators of the full cycle writes 
\begin{equation}
 \mathcal{L}_{rec}^{(f)} = \lVert \vect{I}_r - f_w(\vect{d}'_r, \vect{I}_l) \lVert_1 +  \lVert \vect{I}_l - f_w(\vect{d}'_l, \vect{\hat{I}}_r) \lVert_1
\end{equation}
We add a discriminator $D_l$ for discriminating the synthesized image $\hat{\vect{I}}_l$, and then the adversarial learning strategy is used for both the left and the right image views in a closed loop. The adversarial objective for the full cycled model can be formulated as 
\begin{equation}
\begin{split}
 & \mathcal{L}_{gan}^{(f)}(G_l, G_r, D_r, \vect{I}_{l}, \vect{I}_{r}) =  
  \mathbb{E}_{\vect{I}_r \sim p(\vect{I}_r)}[\log D_r(\vect{I}_r)] \\ +  & \mathbb{E}_{\vect{I}_l \sim p(\vect{I}_l)}[\log(1 - D_r(f_w(\vect{d}'_r, \vect{I}_l)))] +
   \mathbb{E}_{\vect{I}_l \sim p(\vect{I}_l)}[\log D_l(\vect{I}_l)] \\ + & \mathbb{E}_{\vect{I}_r \sim p(\vect{I}_r)}[\log(1 - D_l(f_w(\vect{d}'_l, \vect{\hat{I}}_r)))]
\end{split}
\end{equation}
Each half of the cycle network produces a disparity map corresponding to a different view translation, \ie~$\vect{d}'_l$ and $\vect{d}'_r$. To make them constrain on each other, we add an $L1$-norm consistence loss between these two maps as follows:
\begin{equation}
\mathcal{L}_{con}^{(f)} = || \vect{d}'_l - f_w(\vect{d}'_l, \vect{d}'_r) ||_1
\end{equation}
where since the two disparity maps are for different views and are not aligned, we use the warping operation to make them pixel-to-pixel matched. The consistence loss put a strong view constraint for each half cycle and thus facilitates the learning of both half cycles.
\par\textbf{Full objective.} The full optimization objective consists of the reconstruction losses of both generators, the adversarial losses for both view synthesis and the half-cycle consistence loss. It can be written as follows:
\begin{equation}
\mathcal{L}_{o}^{(f)} = \gamma_1 \mathcal{L}_{rec}^{(f)} + \gamma_2 \mathcal{L}_{gan}^{(f)} + \gamma_3 \mathcal{L}_{con}^{(f)}.
\end{equation}
Where $\{\gamma_i\}_{i=1}^3$ represents a set of weights for controlling the importance of different optimization parts.

\par \textbf{Inference.} When the optimization is finished, given a testing pair $\{\vect{I}_l, \vect{I}_r\}$, the testing is performed by combining the output disparity maps $\vect{d}'_l$ and $\vect{d}'_r$ in a weighted averaging scheme. We treat the two half cycles with equal importance, and the final disparity map $\vect{D}$ is obtained as the mean of the two, \ie~$D = (\vect{d}'_l + f_w(\vect{d}'_l, \vect{d}'_r))/2.$

\subsection{Network Implement Details}
To describe the details of the network implementation, in terms of the generators $G_l$ and $G_r$, we use a ResNet-50 backbone network for the encoder part, and the decoder part contains five deconvolution with ReLU operations in which each 2 times up-samples the feature map. The skip connections are also used to pass information from the backbone representations to the deconvolutional feature maps for obtaining more effective feature aggregation. For the discriminators $D_l$ and $D_r$, we employ the same network structure which has five consecutive convolutional operations with a kernel size of 3, a stride size of 2 and a padding size of 1, and batch normalization~\cite{ioffe2015batch} is performed after each convolutional operation. Adversarial loss is applied to output patches. For the warping operation, a bilinear sampler is used as in~\cite{godard2017unsupervised}. 

%% file: exp.tex
\section{Experimental Results}
\label{sec:exp}

We present both qualitative and quantitative results on publicly available datasets to demonstrate the performance of the proposed approach for unsupervised adversarial depth estimation.

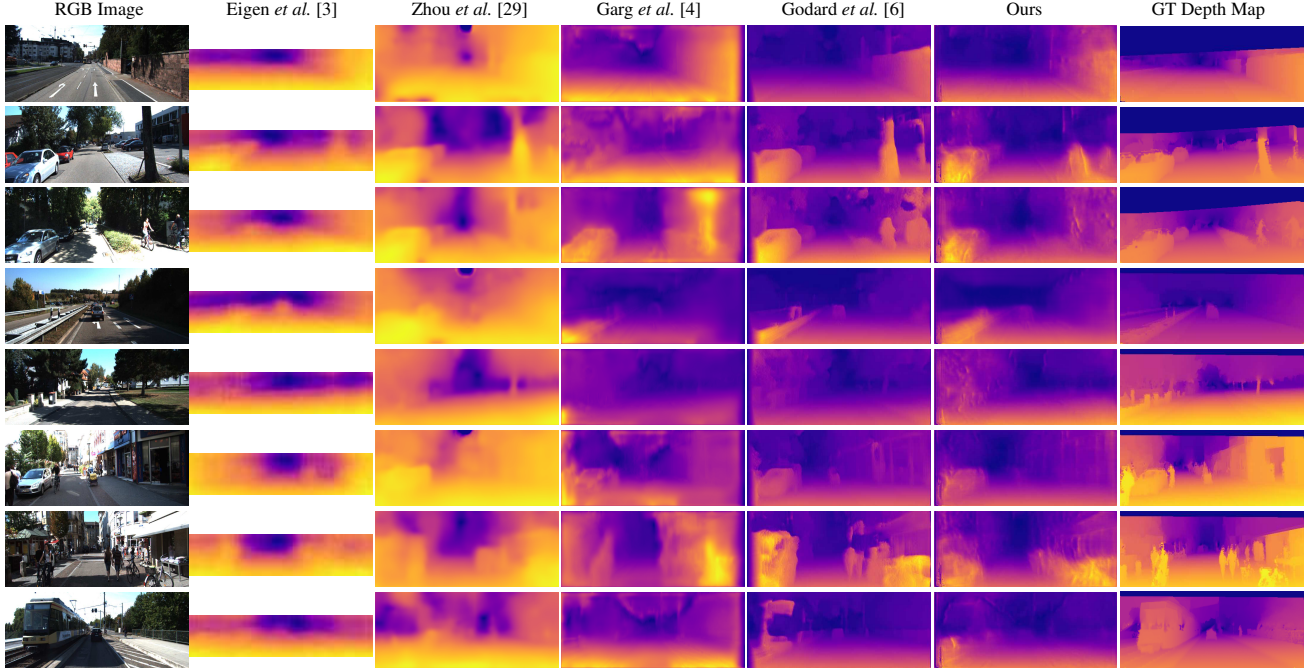
\begin{figure*}[!t]
\centering
\includegraphics[angle=270, width=\textwidth]{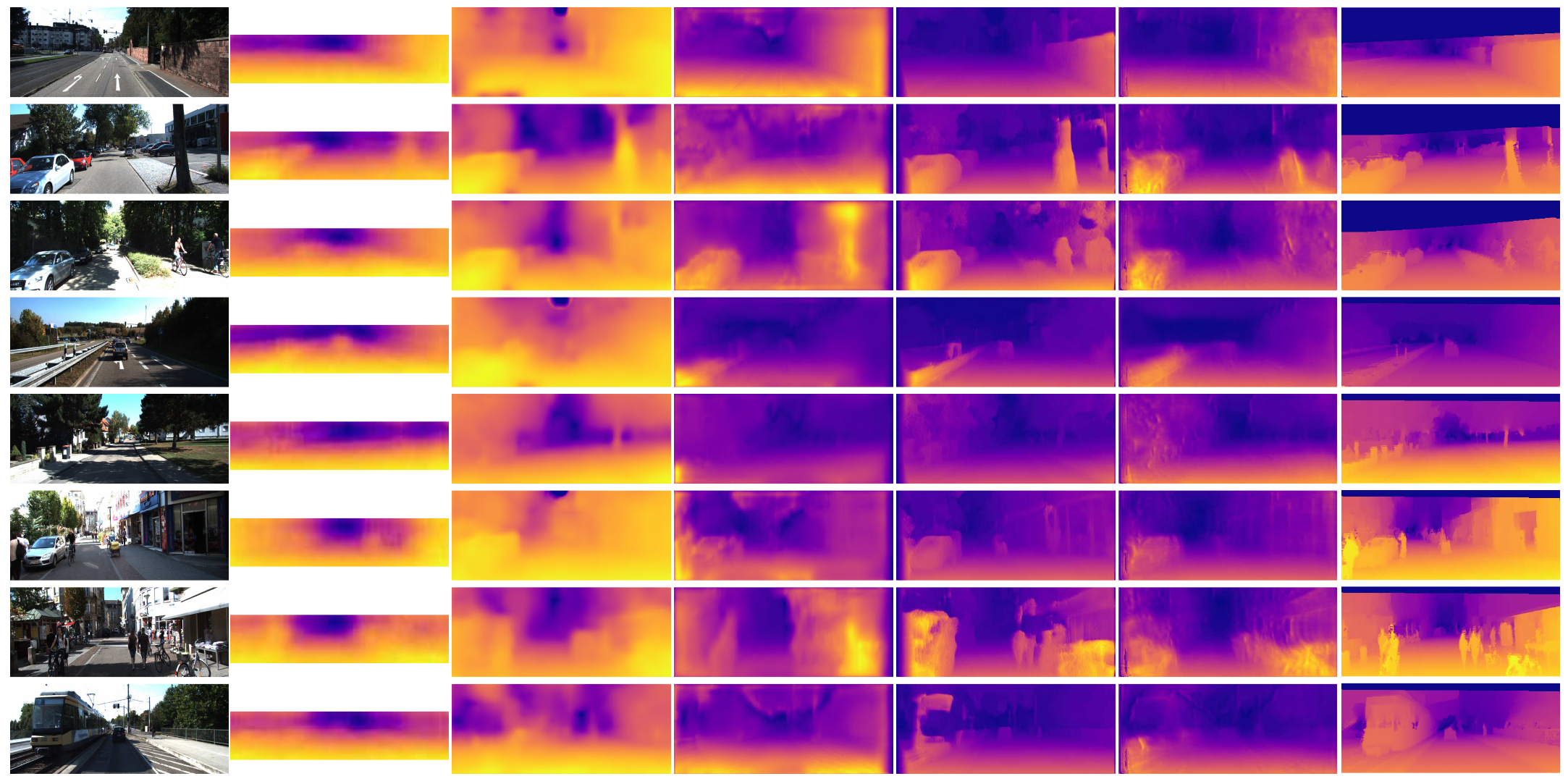}
	\put(-475,2){\scriptsize RGB Image}
	\put(-410,2){\scriptsize Eigen~\etal~\cite{eigen2014depth}}
	\put(-340,2){\scriptsize Zhou~\etal~\cite{zhou2017unsupervised}}
	\put(-270,2){\scriptsize Garg~\etal~\cite{garg2016unsupervised}}
	\put(-200,2){\scriptsize Godard~\etal~\cite{godard2017unsupervised}}
	\put(-115,2){\scriptsize Ours }
	\put(-60,2){\scriptsize GT Depth Map}
 \caption{Qualitative comparison with different competitive approaches with both supervised and unsupervised settings on the KITTI test set. The sparse groundtruth depth maps are filled with bilinear interpolation for better visualization.}
\end{figure*}

\begin{table*}[!t]
\begin{center}
\begin{tabular}{ | l | c || c | c | c | c | c | c | c |}
\toprule
   \multirow{2}{*}{Method} & \multirow{2}{*}{Sup} & Abs Rel & Sq Rel & RMSE & RMSE log & $\delta<1.25$ & $\delta<1.25^2$ & $\delta<1.25^3$ \\
   \cline{3-6} \cline{7-9}
    && \multicolumn{4}{c|}{lower is better}& \multicolumn{3}{c|}{higher is better} \\
\midrule
    Half-Cycle Mono              & N & 0.240 & 4.264 & 8.049 & 0.334 & 0.710 & 0.871 & 0.937 \\
    Half-Cycle Stereo                 & N & 0.228 & 4.277 & 7.646 & 0.318 & 0.748 & 0.892 & 0.945 \\
    Half-Cycle + D                 & N & 0.211 & 2.135 & 6.839 & 0.314 & 0.702 & 0.868 & 0.939\\
    Full-Cycle + D              & N & 0.198 & \textbf{1.990} & \textbf{6.655} & \textbf{0.292} & 0.721 & 0.884 & 0.949\\
    Full-Cycle + D + SE         & N & 0.\textbf{190} & 2.556 & 6.927 & {0.353} & \textbf{0.751} & \textbf{0.895} & \textbf{0.951}\\
\bottomrule
\end{tabular}
\end{center}
\caption{Quantitative evaluation results of different variants of the proposed approach on the KITTI dataset for the ablation study. We do not perform cropping on the dpeth maps for evaluation and depth range is from 0 to 80 meters.}
\label{tab:ablation_kitti}
\vspace{-15pt}
\end{table*}

\subsection{Experimental Setup}
\textbf{Datasets.} We carry out experiments on two large datasets, \ie~KITTI~\cite{kitti} and Cityscapes~\cite{Cityscapes} . For the \textbf{KITTI} dataset, we use the Eigen split~\cite{eigen2014depth} for training and testing. This split contains 22,600 training image pairs, and 697 test pairs. We do data augmentation with online random flipping of the images during training. The \textbf{Cityscapes} dataset is collected using a stereo camera from a driving vehicle through several German cities, during different times of the day and seasons. It presents higher resolution images and is annotated mainly for semantic segmentation. To train our model we combine the densely and coarse annotated splits to obtain 22,973 image-pairs. For testing we use the 1,525 image-pairs of the densely annotated split. The test set also has pre-computed disparity maps for the evaluation. 
\par\textbf{Parameter Setup.} The proposed model is implemented using the deep learning library~\textit{TensorFlow}~\cite{tensorflow2015-whitepaper}. The input images are down-sampled to a resolution of $512 \times 256$ from $1226 \times 370$ in the case of the KITTI dataset, while for the Cityscapes dataset, at the bottom one fifth of the image is cropped following~\cite{godard2017unsupervised} and then is resized to $512 \times 256$. The output disparity maps from two input images are fused with a learned linear combination to obtain the final disparity map with a size $512 \times 256$. The batch size for training is set to 8 and the initial learning rate is $10^{-5}$ in all the experiments. We use the Adam optimizer for the optimization. The momentum parameter and the weight decay are set to $0.9$ and $0.0002$, respectively. The final optimization objective has weighed loss parameters $\gamma_1=1$, $\gamma_2=0.1$ and $\gamma_3=0.1$. The learning rate is reduced by half at both $[80k, 100k]$ steps. For our experiments we used an NVIDIA Tesla K80 with 12 GB of memory.

\textbf{Detailed Training Procedure.} We train the half-cycle model with a standard training procedure, \ie~initializing the network with random weights and making the network train for a full 50 epochs. For the cycled model we optimize the network with an iterative training procedure. After random weights initialization, we train the first half branch $\{\vect{I}_l, \vect{I}_r\} \to \hat{\vect{I}}_r$, with generator $G_l$ and discriminator $D_r$ for a 20k iteration steps. After that we train the second half branch $\{\hat{\vect{I}}_r,\vect{I}_l\} \to \hat{\vect{I}}_l$ with generator $G_r$ and discriminator $D_l$ for another 20k iterations. For the training of the first cycle branch, we do not use the cycle consistence loss since the second half branch is not trained yet. Finally we jointly train the whole network with all the losses embedded for a final round of 100k iterations.

\textbf{Evaluation Metrics.} To quantitatively evaluate the proposed approach, we follow several standard evaluation metrics used in previous works~\cite{eigen2014depth, godard2017unsupervised, wang2015towards}. Given $P$ the total number of pixels in the test set and $\hat{d}_i$, $d_i$ the estimated depth and ground truth depth values for pixel $i$, we have (i) the mean relative error (abs rel): 
$\frac{1}{P} \sum_{i=1}^{P} \frac{\parallel \hat{d}_i - d_i \parallel}{d_i}$, 
(ii) the squared relative error (sq rel): 
$\frac{1}{P} \sum_{i=1}^{P} \frac{\parallel \hat{d}_i - d_i \parallel^2}{d_i}$, 
(iii) the root mean squared error (rmse): 
$\sqrt{\frac{1}{P}\sum_{i=1}^P(\hat{d}_i - d_i)^2}$, 
(iv) the mean $\log10$ error (rmse log):
$\sqrt{\frac{1}{P} \sum_{i=1}^{P} \parallel \log \hat{d}_i - \log d_i \parallel^2}$
(v) the accuracy with threshold $t$, \ie the percentage of $\hat{d}_i$ such that $\delta = \max (\frac{d_i}{\hat{d}_i},\frac{\hat{d}_i}{d_i}) < t$, where $t \in [1.25, 1.25^2, 1.25^3]$.

\subsection{Ablation Study}

\begin{table*}
	\begin{center}
		\begin{tabular}{ | l | c || c | c | c | c | c | c | c |}
			\toprule
			\multirow{2}{*}{Method} & \multirow{2}{*}{Sup} & Abs Rel & Sq Rel & RMSE & RMSE log & $\delta<1.25$ & $\delta<1.25^2$ & $\delta<1.25^3$ \\
			\cline{3-6} \cline{7-9}
			&& \multicolumn{4}{c|}{lower is better}& \multicolumn{3}{c|}{higher is better} \\
			\midrule
			Saxena~\etal~\cite{saxena2006learning}&Y&0.280&-&8.734&-&0.601&0.820&0.926\\
			Eigen~\etal~\cite{eigen2014depth}&Y&0.190&1.515&7.156&0.270&0.692&0.899&0.967\\
			Liu~\etal~\cite{liu2016learning}&Y&0.202&1.614&6.523&0.275&0.678&0.895&0.965\\
			AdaDepth~\cite{adadepth}, 50m&Y&0.162&1.041&4.344&0.225&0.784&0.930&0.974\\
			Kuznietzov~\etal~\cite{kuznietsov2017semi}&Y&-&-&4.815&0.194&0.845&0.957&0.987\\
			Xu~\etal~\cite{xu2018monocular}  & Y & 0.132 & 0.911 & - & 0.162 &0.804 &0.945 & 0.981 \\
			\midrule
			Zhou~\etal~\cite{zhou2017unsupervised}&N&0.208&1.768&6.856&0.283&0.678&0.885&0.957\\
			Garg~\etal~\cite{garg2016unsupervised}&N&0.169&1.08&5.104&0.273&0.740&0.904&0.962\\
			AdaDepth~\cite{adadepth}, 50m&N&0.203&1.734&6.251&0.284&0.687&0.899&0.958\\
			Godard~\etal~\cite{godard2017unsupervised}&N&\textbf{0.148}&\textbf{1.344}&\textbf{5.927}&\textbf{0.247} &\textbf{0.803}&\textbf{0.922}&\textbf{0.964}\\
			\midrule
			Ours, 80m & N & 0.166 & 1.466 & 6.187 & 0.259 & 0.757 & 0.906 & 0.961\\
			Ours with shared enc, 80m & N & {0.152} & {1.388} & {6.016} &{0.247} & {0.789}& {0.918} & {0.965}\\
			Ours, 50m& N & 0.158 & 1.108 & 4.764 & 0.245 & 0.771 & 0.915 & 0.966\\
			Ours with shared enc, 50m             & N &\textbf{0.144} & {\textbf{1.007}} & {\textbf{4.660}} & {\textbf{0.240}} & {\textbf{0.793}} & {\textbf{0.923}} & {\textbf{0.968}}\\
			\bottomrule
		\end{tabular}
	\end{center}
	\caption{Comparison with state of the art. Training and testing are performed on the KITTI \cite{kitti} dataset. Supervised and semi-supervised methods are marked with Y in the supervision column, unsupervised methods with N. Numbers are obtained on Eigen test split with Garg image cropping. Depth predictions are capped at the common threshold of 80 meters, if capped at 50 meters we specify it.}
\label{tab:state-art_kitti}
\end{table*}

\begin{table*}[!t]
\begin{center}
\begin{tabular}{ | l | c || c | c | c | c | c | c | c |}
\toprule
   \multirow{2}{*}{Method} & \multirow{2}{*}{Sup} & Abs Rel & Sq Rel & RMSE & RMSE log & $\delta<1.25$ & $\delta<1.25^2$ & $\delta<1.25^3$ \\
   \cline{3-6} \cline{7-9}
    && \multicolumn{4}{c|}{lower is better}& \multicolumn{3}{c|}{higher is better} \\
\midrule
    Half-Cycle Mono                & N & 0.467 & 7.399 & 5.741 & 0.493 & 0.735 & 0.890 & 0.945 \\
    Half-Cycle Stereo                  & N & 0.462 & 6.097 & 5.740 & 0.377 & 0.708 & 0.873 & 0.937 \\
    Half-Cycle + D                 & N & \textbf{0.438} & \textbf{5.713} & 5.745 & 0.400 & 0.711 & 0.877 & 0.940 \\
    Full-Cycle + D              & N & 0.440 & 6.036 & \textbf{5.443} & \textbf{0.398} & \textbf{0.730} & \textbf{0.887} & \textbf{0.944} \\
\bottomrule
\end{tabular}
\end{center}
\caption{Quantitative evaluation results of different variants of the proposed approach on the Cityscapes dataset for the ablation study.}
\label{tab:ablation_city}
\end{table*}

To validate the adversarial learning strategy is beneficial for the unsupervised depth estimation, and the proposed cycled generative network is effective for the task, we present an extensive ablation study on both the KITTI dataset (see Table~\ref{tab:ablation_kitti}) and on the Cityscape dataset (see Table~\ref{tab:ablation_city}). 

\par\textbf{Baseline Models.} We have several baseline models for the ablation study, including (i) Half-cycle with a monocular setting (half-cycle mono), which uses a straight forward branch to synthesize from one image view to the other with a single disparity map output and the single RGB image is as input during testing; (ii) half-cycle with a stereo setting (half-cycle stereo), which uses a straight forward branch but with two disparity maps produced and combined; (iii) half-cycle with a discriminator (half-cycle + D), which use a single branch as in (ii) while adds a discriminator for the image synthesis; (iv) full-cycle with two discriminators (full-cycle + D), which is our whole model using a full cycle with two discriminators added; (v) full-cycle with two discriminators and sharing encoders (full-cycle + D + SE), which has the same structure as (iv) while the parameters of the encoders of the generators are shared.

\par\textbf{Evaluation on KITTI.} As we can see from Table~\ref{tab:ablation_kitti}, the baseline model Half-Cycle Stereo shows significantly better performance on seven out of eight evaluation metrics than the baseline model Half-Cycle Mono, demonstrating that the utilization of the stereo images and the combination of the two estimated complementary disparity maps clearly boosts the performance.  

\begin{figure*}
	\centering
	\includegraphics[angle=270, width=\textwidth]{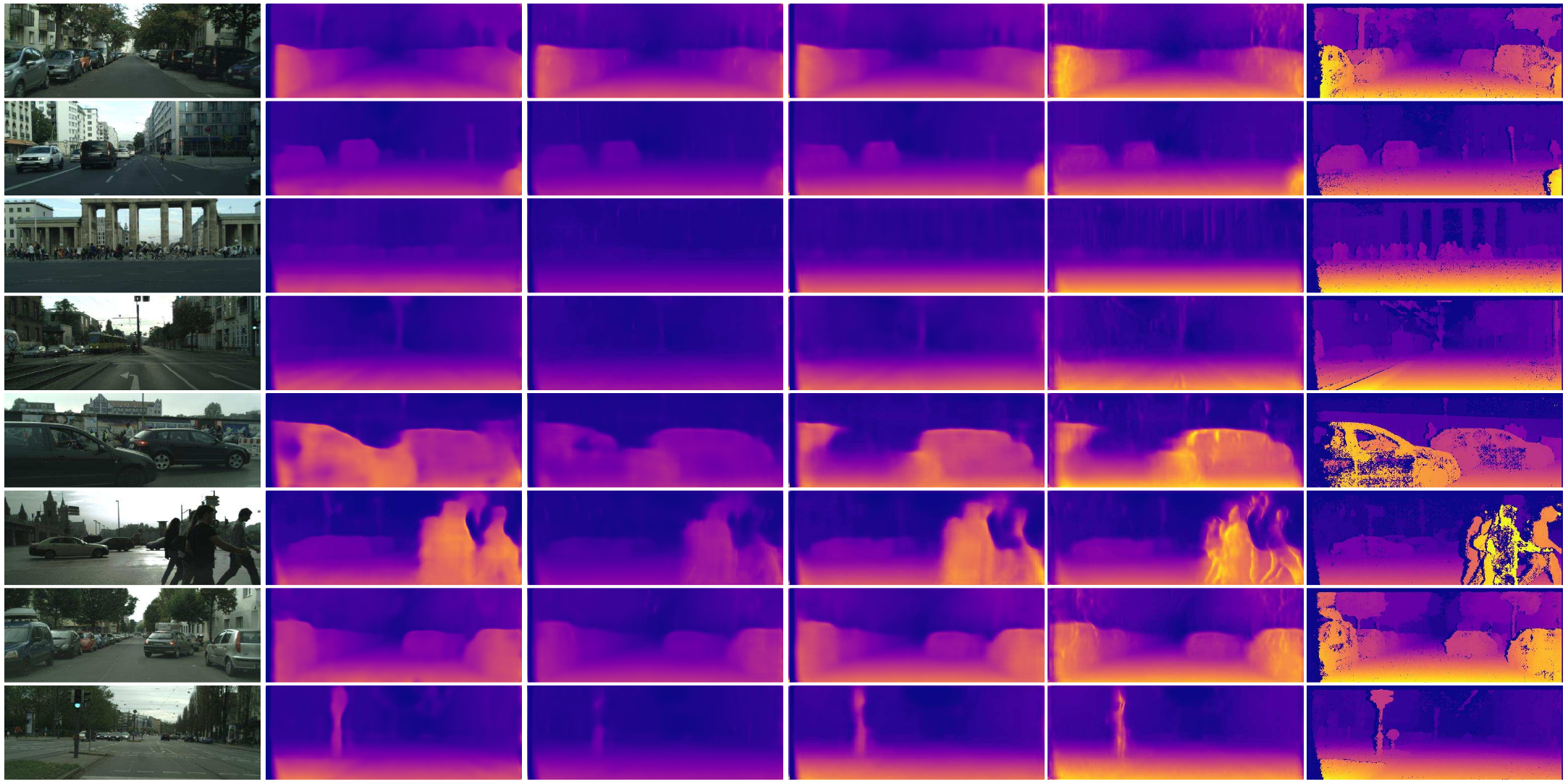}
	\put(-470,2){\scriptsize RGB Image}
	\put(-395,2){\scriptsize Half-Cycle Mono}
	\put(-310,2){\scriptsize Half-Cycle Stereo}
	\put(-230,2){\scriptsize Half-Cycle + D}
	\put(-150,2){\scriptsize Full-Cycle + D}
	\put(-65,2){\scriptsize GT Depth Map}
	\caption{Qualitative comparison of different baseline models of the proposed approach on the Cityscapes testing dataset.}
	\label{fig:cityscapes}
	\vspace{-8pt}
\end{figure*}
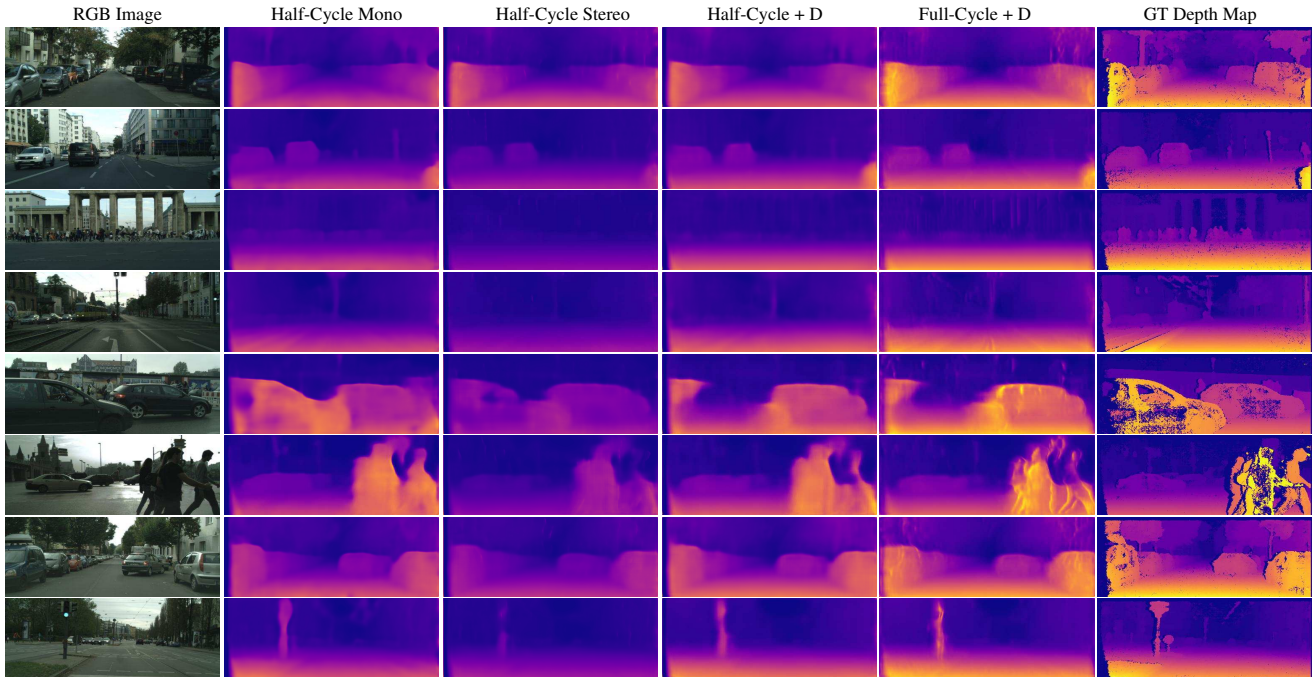

\par By using the adversarial learning strategy for the image synthesis, the baseline Half-Cycle + D outperforms the baseline Half-Cycle Stereo with around 1.7 points gain on the metric of Abs Rel, which verifies our initial intuition of using the adversarial learning to improve the quality of the image synthesis, and thus gain the improvement of the disparity prediction. In addition, we also observe in the training process, the adversarial learning helps to maintain a more stable convergence trend with small oscillations in terms of the training loss than the one without it (\ie~Half-Cycle Stereo), probably leading to a better optimized model. 

\par It is also clear to observe that the proposed cycled generative network with adversarial learning (Full-Cycle + D) achieved much better results than the models with only half cycle (Half-Cycle + D) on all the metrics. Specifically, the Full-Cycle + D model improves the Abs Rel around 2 points, and also improves the accuracy a1 around 1.9 points over Half-Cycle + D. The significant improvement demonstrates the effectiveness of the proposed network design, confirming that the cycled strategy brings stronger constraint and supervision to optimize the both generators. Finally, we also show that the propose cycled model using a sharing encoder for the generator (Full-Cycle + D + SE). By using the sharing structure, we obtain even better results than the non-sharing model (Full-Cycle + D), which is probably because the shared one has a more compact network structure and thus is relatively easier to optimize with a limited number of training samples. 


\textbf{Evaluation on Cityscapes.} We also conduct another ablation study on the Cityscapes dataset and the results are shown in Table~\ref{tab:ablation_city}. We can mostly observe similar trend of the performance gain of the different baseline models as we already analyzed on the KITTI dataset. The performance comparison of the baselines on this challenging dataset further confirms the advantage of the proposed approach. For the comparison of the model Half-Cycle + D and the model Full-Cycle + D, although the latter one achieves slightly worse results on the first two error metrics, it still produces clearly better performance on the rest six evaluation metrics. Since there is no official evaluation protocol for depth estimation on this dataset, the results are evaluated with the protocol on the KITTI, and are directly evaluated on the disparity maps as they are directly proportional to each other. In Fig.~\ref{fig:cityscapes}, some qualitative comparison of the baseline models are presented.

\subsection{State of the Art Comparison}
In Table~\ref{tab:state-art_kitti}, we compare the proposed full model with several state-of-the-art methods, including the ones with the supervised setting, \ie~Saxena~\etal~\cite{saxena2006learning}, Eigen~\etal~\cite{eigen2014depth}, Liu~\etal~\cite{liu2016learning}, AdaDepth~\cite{adadepth}, Kuznietzov~\etal~\cite{kuznietsov2017semi} and Xu~\etal~\cite{xu2018monocular}, and the ones with the unsupervised setting, \ie~Zhou~\etal~\cite{zhou2017unsupervised}, AdaDepth~\cite{adadepth}, Garg~\etal~\cite{garg2016unsupervised} and Godard~\etal~\cite{godard2017unsupervised}. Among all the supervised approaches, we have achieved very competitive performance to the best one of them (\ie~Xu~\etal~\cite{xu2018monocular}), while ours is totally unsupervised without using any ground-truth depth data in training. For comparison with the unsupervised methods, we are also very close to the best competitor (\ie~Godard~\etal~\cite{godard2017unsupervised}). AdaDepth~\cite{adadepth} is the most technically related to our approach, which considers adversarial learning in a context of domain adaptation with extra synthetic training data. Ours significantly outperforms their results with both the supervised and unsupervised setting, further demonstrating the effectiveness of the means we considered and proposed for unsupervised depth estimation with the adversarial learning strategy. As far as we know, there are not quantitative results presented in the existing works on the Cityscapes dataset.
\subsection{Analysis on the Time Aspect.}
For the training of the whole network model, on a single Tesla K80 GPU, it takes around 45 hours on KITTI dataset with around 22k training images. For the running time, in our case with the resolution of $512 \times 256$, the inference of one image takes around 0.140 seconds, which is a near real-time processing speed.